\documentclass[a4paper,headings=small,parskip=half]{scrartcl}

\usepackage{fourier}
\usepackage[british]{babel}
\usepackage[utf8]{inputenc}
\usepackage{natbib}

\setkomafont{sectioning}{\bfseries\normalsize\normalcolor}

\begin{document}
\textbf{How to Write a Bias Statement\\
Recommendations for Submissions to the Workshop on Gender Bias in NLP}

{\itshape
Christian Hardmeier\\
IT University of Copenhagen and Uppsala University\\
chrha@itu.dk

Marta R.\ Costa-jussà\\
Universitat Politècnica de Catalunya\\
marta.ruiz@upc.edu

Kellie Webster\\
Google AI Language\\
websterk@google.com

Will Radford\\
Canva\\
wejradford@gmail.com

Su Lin Blodgett\\
Microsoft Research\\
sulin.blodgett@microsoft.com}
\bigskip

{\itshape At the Workshop on Gender Bias in NLP (GeBNLP), we'd like to encourage
authors to give explicit consideration to the wider aspects of bias and its
social implications. For the 2020 edition of the workshop, we therefore
requested that all authors include an explicit \emph{bias statement} in their
work to clarify how their work relates to the social context in which NLP
systems are used.

The programme committee of the workshops included a number of
reviewers with a background in the humanities and social sciences, in addition
to NLP experts doing the bulk of the reviewing. Each paper was assigned one of
those reviewers, and they were asked to pay specific attention to the provided
bias statements in their reviews. This initiative was well received by the
authors who submitted papers to the workshop, several of whom said they received
useful suggestions and literature hints from the bias reviewers. We are
therefore planning to keep this feature of the review process in future editions
of the workshop.

This document was originally published as a blog post on the
web site of GeBNLP 2020.}

\section{Introduction}

The idea behind the requirement of a bias statement is to encourage a common
format for discussing the assumptions and normative stances inherent in any
research on bias, and to make them explicit so they can be discussed. This is
inspired by the recommendations by Blodgett et al. (2020), and we borrow from
them in our definition of the bias statement. In this document, we provide some
guidance to help you write a bias statement for your research.

Two things are worth highlighting. Firstly, this document is intended to help
you write a bias statement, but perhaps your work is a bit different from what
we had in mind when we wrote it. That's fine -- we're really keen on promoting
this discussion. Although we don't require a specific form of the statement, we
suggest you make a specific section for this statement. If your case is
different, do whatever makes sense. The reviewers will be asked to comment on
your bias statement in the specific context of your work, and we've recruited
some reviewers from the social sciences and humanities to help us with that.
And secondly, we'd like to encourage you to think about your concepts of bias
and how they relate to the lived experience of humans throughout your work,
from the beginning to the end. That's the really important thing. The bias
statement is just a way to condense the discussion in one place.

\section{Types of Harm}

One part of a successful bias statement is to clarify \emph{what type of harm}
we are worried about, and \emph{who suffers} because of it. Doing so explicitly
serves two purposes. On the one hand, by describing certain behaviours as
harmful, we make a judgement based on the values we hold. It's a normative
judgement, because we declare that one thing is right (for instance, treating
all humans equally), and another thing wrong (for instance, exploiting humans
for profit). On the other hand, being explicit about our normative assumptions
also makes it easier to evaluate, for ourselves, our readers and reviewers,
whether the methods we propose are in fact effective at reducing the harmful
effects we fear, and that will help us make progress more quickly.

This schema is not final. We don't necessarily request that you adhere strictly
to these categories, but we'd like to highlight these these broad categories to
get your imagination going about what kinds of harms might arise.

Following the categories of \citet{blodgett-etal-2020-language}:

\textbf{Allocational harms:} An automated system allocates resources or
opportunities unfairly to different social groups.

\begin{addmargin}{2em}
\textit{Example:}\\
A recruitment support system based on a database that is trained only on men
systematically ranks women lower when they are compared with similarly qualified
men.
\end{addmargin}

\textbf{Representational harms}  arise when a system represents some social
groups in a less favorable light than others, demeans them, or fails to
recognize their existence altogether.

\begin{addmargin}{2em}
\textit{Examples:}

\textit{Stereotyping:} Propagating negative generalisations about particular social groups

\textit{Differences in system performances affecting users unequally:} language
that misrepresents the distribution of social groups or language that
denigrates certain social groups
\end{addmargin}

\section{Recommendations for authors}

\begin{itemize}
\item   Provide explicit statements of why the system behaviours described as
“bias” are harmful, in what ways, and to whom. Authors should reflect critically
over their own definitions of ``bias'' -- Are there any limitations to choosing this
definition, any injustices or undesirable outcomes that it might overlook that
might require a different kind of definition, any implicit assumptions this
definition makes or requires that might not always hold true? This is
essentially the same kind of discussion that any analysis of modeling decisions
generally requires.

\item   Be forthright about the normative reasoning underlying these statements.

        \emph{Negative example:} ``Biased word embeddings can lead to biased downstream systems and contribute to social injustice.''

        \emph{Positive examples:} ``Coreference systems with gender labels that treat gender as fixed, immutable, and binary are harmful because they erase or exclude non-binary or transgender people''; or ``Toxicity systems that treat Mainstream U.S. English as more toxic than African-American English are harmful because they contribute to the stigmatization of African-American English, may disenfranchise AAE speakers online, and may result in burdens of dealing with toxicity systems that are differentially distributed across speaker groups.''
\end{itemize}

\section{A Concrete Example of a Bias Statement}

PAPER: Basta, C., Costa-jussà, M.~R. and Casas, N. Evaluating the Underlying
Gender Bias in Contextualized Word Embeddings. \textit{Proceedings of
the 1st ACL Workshop on Gender Bias for Natural Language Processing}, Florence,
2020.

BIAS STATEMENT: In this paper, we study stereotypical associations between male
and female gender and professional occupations in contextual word embeddings. If
a system systematically and by default associates certain professions with a
specific gender, this creates a representational harm by perpetuating
inappropriate stereotypes about what activities men and women are able, allowed
or expected to perform, e.g. making that there is a pay gap in professionals in
STEM predicted by the confidence gap \citep{Sterling30303}. When such representations are used in downstream NLP
applications, there is an additional risk of unequal performance across genders
\citep{gonen-webster-2020-automatically}. Our work is based on the belief that
the observed correlations between genders and occupations in word embeddings are
a symptom of an inadequate training process, and decorrelating genders and
occupations would enable systems to counteract rather than reinforce existing
gender imbalances

\clearpage

\section*{Acknowledgements}

Christian Hardmeier was supported by the Swedish Research Council under grant
2017-930 and Marta R. Costa-jussà by the European Research Council (ERC) under
the European Union’s Horizon 2020 research and innovation programme (grant
agreement No. 947657). 

\bibliographystyle{apalike}
\bibliography{bias-statement}

\end{document}